\documentclass[runningheads]{llncs}

\usepackage[T1]{fontenc}
\def\doi#1{\href{https://doi.org/\detokenize{#1}}{\url{https://doi.org/\detokenize{#1}}}}
\usepackage{graphicx}
\usepackage{listings}
\usepackage[misc,geometry]{ifsym}
\usepackage{tabularx}
\usepackage{multirow}
\usepackage{amsmath}
\usepackage{xcolor}
\usepackage{blkarray, bigstrut}
\usepackage{hyperref}
\hypersetup{colorlinks=true, linkcolor=blue, citecolor=blue}

\usepackage{amsfonts}
\usepackage{dsfont}
\usepackage{lipsum}
\usepackage{adjustbox}
\usepackage{float}
\usepackage{booktabs}
\usepackage{cite}
\usepackage{bm}
\usepackage{pifont}
\newcommand{\cmark}{\ding{51}}%
\newcommand{\xmark}{\ding{55}}%
\begin{document}
\title{Anatomy-Guided Weakly-Supervised Abnormality Localization in Chest X-rays}
\titlerunning{Anatomy-Guided Weakly-Supervised Abnormality Localization in CXR}
%
\author{Ke Yu\inst{1}{$^{(\textrm{\Letter})}$}\and
Shantanu Ghosh\inst{1} \and
Zhexiong Liu\inst{1}\and
Christopher Deible\inst{2}\and \\
Kayhan Batmanghelich\inst{1}
}
%
\authorrunning{K. Yu et al.}
%
\institute{University of Pittsburgh, Pittsburgh, PA, USA \\
\email{yu.ke@pitt.edu}\\
\and
University of Pittsburgh Medical Center, Pittsburgh, PA, USA
\
}
\maketitle              
\begin{abstract}
Creating a large-scale dataset of abnormality annotation on medical images is a labor-intensive and costly task. Leveraging \emph{weak supervision} from readily available data such as radiology reports can compensate lack of large-scale data for anomaly detection methods. However, most of the current methods only use image-level pathological observations, failing to utilize the relevant \emph{anatomy mentions} in reports. Furthermore, Natural Language Processing (NLP)-mined weak labels are noisy due to label sparsity and linguistic ambiguity. We propose an Anatomy-Guided chest X-ray Network (AGXNet) to address these issues of weak annotation. Our framework consists of a cascade of two networks, one responsible for identifying anatomical abnormalities and the second responsible for pathological observations. The critical component in our framework is an anatomy-guided attention module that aids the downstream observation network in focusing on the relevant anatomical regions generated by the anatomy network. We use Positive Unlabeled (PU) learning to account for the fact that lack of mention does not necessarily mean a negative label. Our quantitative and qualitative results on the MIMIC-CXR dataset demonstrate the effectiveness of AGXNet in disease and anatomical abnormality localization. Experiments on the NIH Chest X-ray dataset show that the learned feature representations are transferable and can achieve the state-of-the-art performances in disease classification and competitive disease localization results. Our code is available at \url{https://github.com/batmanlab/AGXNet}.

\end{abstract}
\section{Introduction}
There is considerable interest in developing automated abnormality detection systems for chest X-rays (CXR) in order to improve radiologists' workflow efficiency and reduce observational oversights~\cite{qin2018computer, castellino2005computer, lakhani2017deep, gromet2008comparison}. Typically, training a high-precision detection model using deep learning requires high-quality annotations. However, collecting large-scale annotations by clinical experts is time-consuming and prohibitively expensive. This motivates weakly-supervised learning (WSL) methods~\cite{wang2017chestx, liu2019align, li2018thoracic, bhalodia2021improving} that leverage weak supervision from paired CXR reports that are readily available on a  large scale. There are, however, two unaddressed challenges: (1) The entire report is often summarized to a small set of disease labels, which misses the opportunity of incorporating anatomical context, and (2) Weak labels derived from radiology reports using Natural Language Processing (NLP) are noisy due to the sparsity of the labels and the ambiguity of the language. Improper handling of labeling noise can result in underperforming models. We propose a novel WSL framework to bridge these two gaps.

\begin{figure}[t]
\centering
\includegraphics[width=0.65\textwidth]{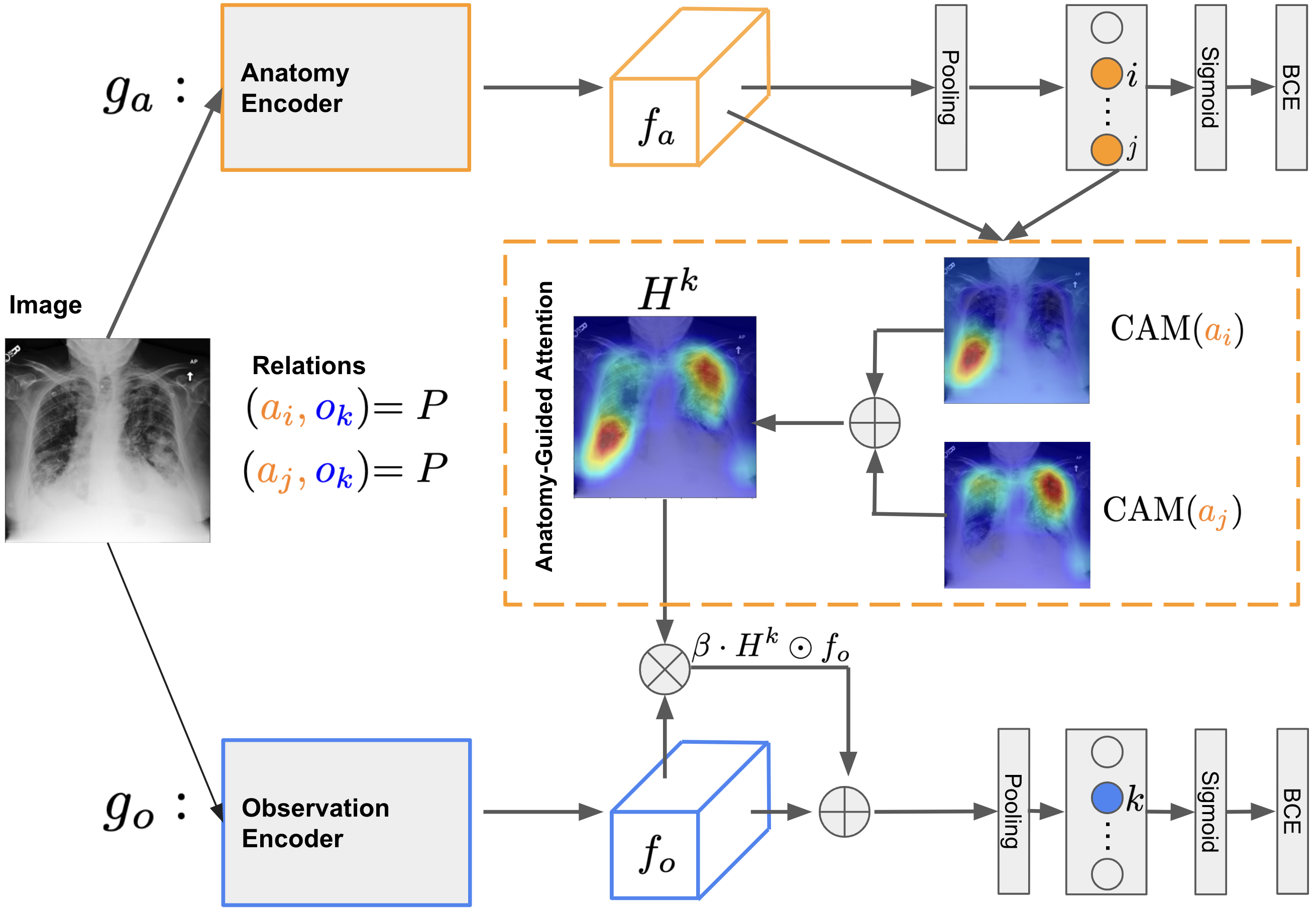}
\caption{ Schematic diagram of the proposed AGXNet. Our architecture comprises of two classification networks, $g_a$ for anatomical abnormalities and $g_o$ for pathological observations. Relations $(a_i, o_k)=P$, $(a_j,o_k)=P$ are parsed from the report and represent that the observation $o_k$ is annotated as \textit{Present} in two anatomical landmarks $a_i, a_j$, respectively. We obtain the anatomy-guided attention map $H^k$ by aggregating CAMs of $a_i, a_j$ and incorporate the $H^k$ as a residual attention into $g_o$. The symbols $f_a, f_o$ denote the intermediate anatomy and observation features, respectively. 
}
\label{fig:2}
\end{figure}

A variety of automated text labelers~\cite{irvin2019chexpert, wang2017chestx, peng2018negbio, jain2021visualchexbert} have been developed to extract image-level observations from CXR reports. However, they do not consider \emph{anatomy mentions} that provide important contexts for associated observations. Aiming to fill this gap, the recently developed pipelines, Chest ImaGenome~\cite{wu2021chest} and RadGraph~\cite{jain2021radgraph} extract fine-grained \textit{``observation-located at-anatomy''} relations, (e.g., ``\textit{opacity in the right lower lung}''), from CXR reports. Yet none of them has been incorporated into a weakly-supervised disease detection method. We build our WSL framework based on the RadGraph dataset.
Radiologists typically employ a systematic approach~\cite{sait2021teaching} when reading CXR images to ensure that no significant abnormality is missed. This approach is essentially documented in CXR reports which typically have imaging observations paired with the related anatomical locations. We design an architecture that reflects this process.

The second challenge is that weak labels extracted from CXR reports are inherently noisy. Typically, only a subset of weak labels is mentioned in a report, and unlabeled data is handled using a basic zero-imputation strategy. However, lack of mention does not necessarily mean a negative label. For example, a CXR report that contains mentions of strong visual clues for pneumonia, such as \textit{opacity} or \textit{consolidation}, may not specifically establish \textit{pneumonia} as a diagnosis due to human variability in the reporting process. We also consider uncertainty mentions (e.g., \textit{may, possible, can't exclude}) as unlabeled, where the noise originates from the intrinsic ambiguities in reports. We formulate this problem as a Positive Unlabeled (PU) learning~\cite{bekker2020learning} where the learner has access to a small set of examples with high-confidence labels (either positive or negative) and a large amount of unlabeled data mixed by positive and negative examples with an unknown mixture ratio. 

In this paper, we present Anatomy-Guided chest X-ray Networks (AGXNet), a novel WSL framework that leverages information from pathological observations and their associated anatomical landmarks mentioned in CXR reports. Our architecture consists of a cascade of two networks, with the upstream anatomy network tasked with identifying anatomical abnormalities in CXR images and the downstream observation network tasked with identifying pathological observations. The inclusion of an anatomy-guided attention (AGA) module aims to aid the observation network paying attention to abnormalities in the context of associated anatomical landmarks. During training, the AGA module also provides a feedback loop to the upstream anatomy network, thus mutually improving the quality of both types of features. In addition, we adopt a PU learning technique to estimate the fraction of positive samples in the unlabeled data and use a self-training approach to iteratively reduce noise in labels. Our model is trained end-to-end. We evaluate the proposed framework on the MIMIC-CXR~\cite{johnson2019mimic} dataset. Results show that the proposed AGXNet model outperforms both supervised and weakly-supervised baselines in disease localization. In-depth ablation studies demonstrate that both AGA module and PU learning can help to improve localization accuracy. We further evaluate the pre-trained encoders on the NIH Chest X-ray dataset~\cite{wang2017chestx} and show that the transferred models achieve the state-of-the-art (SOTA) results in disease classification and competitive performance in disease localization. 
\label{sec:intro}

\section{Methods}
We parse the report into an adjacency matrix that encodes the relations between the observations and anatomical landmarks. The AGA module connects two networks that are used to predict the presence of observations and anatomical abnormalities in CXR images. We use PU learning to explicitly address noise in unlabeled observations. The proposed framework is illustrated in Fig.~\ref{fig:2}.

\noindent \textbf{Report Representation.} We use the RadGraph~\cite{jain2021radgraph} to parse reports to anatomical landmark tokens $\{a_j\}$, observation tokens $\{o_k\}$, and encode their relations using an adjacency matrix $A$. An entry $A(j,k)$ can take one of the three variants $P, N,$ or $U$, representing whether the observation $o_k$ is \textit{Present, Absent} or \textit{Unlabeled} in the anatomical landmark $a_j$. Note, for $o_k$ that is mentioned without a specified location (e.g., ``\textit{No evidence of pneumothorax}''), we link it to a special anatomical landmark token named \textit{unspecified}. Fig. A.1 shows an example adjacent matrix. We summarize the image-level labels for observation and anatomical landmark using the adjacency matrix. 
An observation $o_k$ is assigned a positive label ($y(o_k)=1$) if column $A(\cdot, k)$ has at least one $P$ entry, a negative label ($y(o_k)=0$) if column $A(\cdot, k)$ has no $P$ entry and at least one $N$ entry, and assigned as unlabeled ($y(o_k)=u$) if column $A(\cdot, k)$ only has $U$ entries. For anatomy, an anatomical landmark is labeled as abnormal $y(a_j)=1$ if row $A(j,\cdot)$ has at least one $P$ entry, otherwise it is labeled as normal $y(a_j)=0$.

\noindent \textbf{Anatomy Network.} 
The anatomy network $g_a$ is responsible for identifying abnormalities from anatomical landmarks. The weighted binary cross-entropy (BCE) loss for $a_j$ is given by:

\begin{equation}
    \mathcal{L}_{a_{j}} (\theta_a, w_a^j) = - b_j^{+} \sum_{y(a_j)=1} \text{ln}(g_a(x)) - b_j^{-} \sum_{y(a_j)=0} \text{ln}(1-g_a(x)),
\end{equation}
where $\theta_a$ is the parameter of anatomy encoder network, $w_{a}^{j}$ is the classification weight corresponding to $a_j$, $x$ is an image, $g_a(x)$ is the predicted probability of being abnormal in $a_j$, $b_j^{+}$ and $b_j^{-}$ are balancing factors introduced in~\cite{rajpurkar2017chexnet}. 

\noindent \textbf{AGA.} We introduce the AGA module to guide the downstream observation network $g_o$ focusing on the relevant anatomical regions mentioned in the radiology report. In detail, we construct the AGA map $H^k$ for observation $o_k$ by aggregating the class activation maps (CAMs)~\cite{zhou2016learning} for locations $\{a_j\}$ where $o_k$ was positively observed, i.e., $A(a_j, o_k)=P$. Formally,

\begin{equation}
    H^{k}(\theta_a, w_a^j) = \sum_{j}\mathds{1} \{ A(a_j, o_k)=P\} \underbrace{\sum_{c} w_{a}^{j}f_a(:,:,c;\theta_a)}_{\text{CAM}(a_j)},
\end{equation}
where $f_a(:,:,c)$ represents the activation of channel $c$ in anatomy feature map $f_a$, and $\mathds{1} \{ A(a_j, o_k)=P\}$ is an indicator function. We then transform the values of $H^k$ to $[0,1]$ using the min-max normalization.

\noindent \textbf{Observation Network.} The observation network $g_o$ is responsible for identifying the presence of pathological observations in CXR images. We incorporate the $H^k$ into $g_o$ as a residual attention map and modify the observation feature map $f_o$ as follows, $f'_{o_{k}} = (1 + \beta \cdot H^k) \odot f_o$, where $\odot$ indicates element-wise multiplication for spatial positions, and $\beta$ is a scaling hyperparameter. The weighted BCE loss for $o_k$ is given by:
\begin{equation}
    \mathcal{L}_{o_{k}}(\mathbf{\Theta}) = - b_k^{+} \sum_{y(o_k)=1} \text{ln}(g_o(x, H^k)) - b_k^{-} \sum_{y(o_k) \in \{0, u\}} \text{ln}(1-g_o(x, H^k)),
\end{equation}
where $\mathbf{\Theta}=(\theta_o, w_o^k, \theta_a, w_a^j)$, $\theta_o$ is the parameter of observation encoder network, $w_{o}^{k}$ is the classification weight corresponding to $o_k$, $g_o(x, H^k)$ is the predicted probability of $o_k$ being present in the image, $b_k^{+}$ and $b_k^{+}$ are balancing factors. It is important to note that mapping all unlabeled examples to negative ($| y(o_k) \in \{0, u\}| = |N|+|U|$) would overlook the noise in the unlabeled data, which can degrade model performance. We explicitly handle the randomness present in unlabeled data and formulate this problem using a PU learning approach.

\noindent \textbf{PU Learning.} The distribution of unlabeled data $P_u$ can be decomposed as, $P_u = \alpha P_p + (1-\alpha)P_n$, where $\alpha$ denotes the mixture proportion of positive examples in the unlabeled data, and $P_p, P_n$ denotes the class-conditional distribution for positive and negative class, respectively. We adopt the method \textit{Best Bin Estimation} (BBE) proposed by Garg \textit{et al.}~\cite{garg2021mixture} to estimate $\alpha$. In short, let $X_p, X_n, X_u$ denote the positive, negative and unlabeled samples for $o_k$ in the validation data set, $\hat{F}_p(z), \hat{F}_u(z)$ denote the empirical cumulative distributions of the predicted probabilities of the observation network, named $Z_p, Z_u$. The mixture proportion $\alpha$ is estimated by minimizing the upper confidence bound of the ratio, namely $(1-\hat{F}_u(z)) / (1-\hat{F}_p(z))$. We integrate the BBE with an iterative self-training approach summarized as follows: (1) warm-start training with treating all unlabeled samples as negative, (2) estimating $\alpha$ using BBE, (3) removing $\alpha$ fraction of unlabeled training samples scored as most positive and relabeling the rest $ 1-\alpha$ unlabeled samples as negative, (4) updating the model using positive samples ($|P|$) and provisional negative samples ($|N| + (1-\alpha)|U|$). We repeat steps 2 to 4 until the classifiers reach the best validation performance.

\noindent \textbf{Optimization.} The final loss is given by: $\mathcal{L} = \frac{1}{N_a}\sum_j^{N_a} \mathcal{L}_{a_j} + \frac{1}{N_o}\sum_k ^{N_o} \mathcal{L}_{o_k}$, where $N_a$, $N_o$ are the number of anatomy and observation labels used in training, respectively. The network is trained end-to-end. Importantly, during training, the gradients $\partial \mathcal{L}_{o_k} / \partial \theta$, $\partial \mathcal{L}_{o_k} / \partial w_a^j$ provide a feedback loop to the anatomy network through the AGA module, mutually reinforcing the anatomy and observation features. During inference time, we do not require any text input and set $H^k=0$.
\label{sec:method}

\section{Experiments}
Experiments are carried out to evaluate the performance of AGXNet in abnormality classification and localization. We conduct ablation studies to validate the efficiency of the AGA module and PU learning . We also test the robustness and transferability of the learned anatomy and observation features.

\noindent \textbf{Experimental Details.} We first evaluate the proposed AGXNet on the MIMIC-CXR dataset~\cite{johnson2019mimic}. The RadGraph's inference dataset~\cite{jain2021radgraph} provides annotations automatically generated by DYGIE++~\cite{wadden2019entity} model for 220,763 MIMIC-CXR reports. We select the corresponding  220,763 frontal images from the MIMIC-CXR dataset and obtain their adjacency matrix representations from the RadGraph annotations. The adjacency matrix of each sample has 48 rows (46 anatomical landmarks, 1 \textit{unspecified}, 1 \textit{other anatomies} representing anatomy tokens in the tail distribution) and 64 columns (63 mostly mentioned observations and 1 \textit{other observations} representing observation tokens in the tail distribution). Appx. B provides additional details of anatomical labels. 
For evaluation of disease localization, we use a held out set
~\cite{tam2020weakly} of MIMIC-CXR images with 390 bounding boxes (BBox) for pneumonia ($196/390$) and pneumothorax ($194/390$) annotated by board-certified radiologists. For evaluation of anatomical abnormality localization, we utilize the anatomy BBox from the Chest ImaGenome dataset
~\cite{wu2021chest}, which were extracted by an atlas-based detection pipeline. We use a 80\%-10\%-10\% train-validation-test split with no patient shared across splits. For both anatomy and observation encoders in AGXNet, we use DenseNet-121~\cite{huang2017densely} with pre-trained weights from ImageNet~\cite{deng2009imagenet} as the backbone. We train our framework to predict presence of abnormality in the 46 anatomical landmarks and the presence of two diseases (i.e., pneumonia and pneumothorax) that have ground truth BBox in~\cite{tam2020weakly}. The $\beta$ is set to be $0.1$ based on validation results. We optimize the networks using SGD with $\text{momentum}=0.9$, $\text{weight decay}=10e^{-4}$  and stop training once the validation error reaches minimum. The learning rate is set to be 0.01 and divided by 10 every 6 epochs. We resize all images to $512 \times 512$ without any data augmentation and set the batch size as 16. The model is implemented in PyTorch and trained on a single NVIDIA GPU with 32G of memory.

\noindent \textbf{Evaluation Metric.} We produce disease-specific CAMs for pneumonia and pneumothorax, and anatomy-specific CAMs for the anatomical landmarks. We apply a thresholding-based bounding box generation method and extract isolated regions in which pixel values are greater than the 95\% quantile of the CAM's pixel value distribution. We evaluate the generated boxes against the ground truth BBox using intersection over union ratio (IoU). A generated box is considered as a true positive when $\text{IoU} > T(\text{IoU})$, where $T(\ast)$ is a threshold. 

\begin{table*}[h]
\caption{Disease localization on MIMIC-CXR. Results are reported as the average over 5 independent runs with standard deviations. The highest values are highlighted in bold, and the best results without BBox annotation are underlined.}
\begin{adjustbox}{max width=0.95\textwidth, center}
\begin{tabular}{lcccccccc}
\toprule
\multicolumn{1}{c}{} &
  \multicolumn{1}{c}{} &
  \multicolumn{1}{c}{} &
  \multicolumn{2}{c}{\textbf{IoU @ 0.1}} &
  \multicolumn{2}{c}{\textbf{IoU @ 0.25}} &
  \multicolumn{2}{c}{\textbf{IoU @ 0.5}} \\ 
  \cmidrule(lr){4-5}
  \cmidrule(lr){6-7}
  \cmidrule(lr){8-9}
\multicolumn{1}{l}{\multirow{-2}{*}{\textbf{Disease}}} &
  \multicolumn{1}{c}{\multirow{-2}{*}{\textbf{Model}}} &
  \multicolumn{1}{c}{\multirow{-2}{*}{\textbf{Supervised}}} &
  \multicolumn{1}{c}{\scriptsize Recall} &
  \multicolumn{1}{c}{\scriptsize Precision} &
  \multicolumn{1}{c}{\scriptsize Recall} &
  \multicolumn{1}{c}{\scriptsize Precision} &
  \multicolumn{1}{c}{\scriptsize Recall} &
  \multicolumn{1}{c}{\scriptsize Precision} \\ 
  \midrule
 &
  RetinaNet~\cite{lin2017focal} &
  \cmark &
  $0.52_{\pm 0.04}$ &
  $0.19_{\pm 0.00}$ &
  $0.41_{\pm 0.04}$ &
  $0.15_{\pm 0.00}$ &
  $\bm{0.26_{\pm 0.00}}$ &
  $0.09_{\pm 0.00}$ \\
 &
  CheXpert~\cite{irvin2019chexpert} &
  \xmark &
  $0.68_{\pm 0.01}$ &
  $0.40_{\pm 0.01}$ &
  $0.47_{\pm 0.01}$ &
  $0.28_{\pm 0.00}$ &
  $0.14_{\pm 0.01}$ &
  $0.08_{\pm 0.01}$ \\
 &
  AGXNet w/o AGA &
  \xmark &
  $0.67_{\pm 0.04}$ &
  $0.41_{\pm 0.03}$ &
  $0.54_{\pm 0.04}$ &
  $0.33_{\pm 0.03}$ &
  \underline{$0.16_{\pm 0.02}$} &
  \underline{$\bm{0.10_{\pm 0.01}}$} \\
 &
  AGXNet w/ AGA &
  \xmark & 
  \underline{$\bm{0.75_{\pm 0.01}}$} &
  \underline{$\bm{0.44_{\pm 0.01}}$} &
  \underline{$\bm{0.60_{\pm 0.02}}$} &
  \underline{$\bm{0.35_{\pm 0.01}}$} &
  \underline{$0.16_{\pm 0.00}$} &
  $0.09_{\pm 0.00}$\\
\multirow{-5}{*}{Pneumothorax} &
  AGXNet w/ AGA + PU &
  \xmark & 
  $0.74_{\pm 0.03}$ &
  $0.43_{\pm 0.02}$ &
  $0.58_{\pm 0.03}$ &
  $0.34_{\pm 0.02}$ &
  $0.13_{\pm 0.01}$ &
  $0.07_{\pm 0.01}$ \\
  \midrule
 &
  RetinaNet~\cite{lin2017focal} &
  \cmark & 
  $0.66_{\pm 0.04}$ &
  $0.27_{\pm 0.04}$ &
  $0.62_{\pm 0.04}$ &
  $0.26_{\pm 0.09}$ &
  $\bm{0.42_{\pm 0.04}}$ &
  $\bm{0.17_{\pm 0.03}}$ \\
 &
  CheXpert~\cite{irvin2019chexpert}&
  \xmark & 
  $0.73_{\pm 0.01}$ &
  $0.40_{\pm 0.01}$ &
  $0.51_{\pm 0.01}$ &
  $0.28_{\pm 0.01}$ &
  $0.19_{\pm 0.01}$ &
  $0.10_{\pm 0.01}$ \\
 &
  AGXNet w/o AGA &
  \xmark &
  $0.69_{\pm 0.01}$ &
  $0.44_{\pm 0.01}$ &
  $0.59_{\pm 0.02}$ &
  $0.37_{\pm 0.02}$ &
  $0.18_{\pm 0.02}$ &
  $0.11_{\pm 0.01}$ \\
 &
  AGXNet w/ AGA &
  \xmark & 
  $0.67_{\pm 0.02}$ &
  $0.46_{\pm 0.01}$ &
  $0.58_{\pm 0.02}$ &
  $0.39_{\pm 0.02}$ &
  $0.18_{\pm 0.03}$ &
  $0.12_{\pm 0.02}$ \\
\multirow{-5}{*}{\footnotesize Pneumonia} &
  AGXNet w/ AGA + PU &
  \xmark & 
  \underline{$\bm{0.73_{\pm 0.01}}$} &
  \underline{$\bm{0.47_{\pm 0.01}}$} &
  \underline{$\bm{0.62_{\pm 0.01}}$} &
  \underline{$\bm{0.40_{\pm 0.01}}$} &
  \underline{$0.20_{\pm 0.01}$} &
  \underline{$0.13_{\pm 0.01}$} \\ 
  \bottomrule
\end{tabular}
\end{adjustbox}
\label{table:1}
\end{table*}

\noindent \textbf{Evaluating Disease Localization.} We trained a RetinaNet~\cite{lin2017focal} using the annotated BBox from~\cite{tam2020weakly} as the supervised baseline and a DenseNet-121 using CheXpert~\cite{irvin2019chexpert} disease labels for CAM-based localization as the WSL baseline. We investigated three variants of AGXNet to understand the effects of AGA module and PU learning. Tab.~\ref{table:1} shows that AGXNet trained with PU learning achieved the best localization results for pneumonia at $T(\text{IoU}) = 0.1$ and $0.25$, while AGXNet trained without PU learning performed best in pneumothorax localization at the same IoU thresholds. Note that the label noise in pneumothorax is intrinsically low, thus adding PU learning may not significantly improve its localization accuracy. RetinaNet achieved the best localization results at $T(\text{IoU}) = 0.5$, probably due to its direct predictions for the coordinates of BBox. 
\begin{table*}[t]
\caption{Ablation studies on AGA and PU Learning. Results are averaged over 5 runs. $\alpha$: mixture proportion of positive samples. AA: anatomic abnormality. LAL: left apical lung. LL: left lung. RAL: right apical lung. RLL: right lower lung. RL: right lung.}
\centering
\begin{adjustbox}{max width=0.75\textwidth, center}
\begin{tabular}{cccccccccc}
\toprule
\multicolumn{1}{c}{\multirow{3}{*}{\textbf{Model}}} &
  \multicolumn{4}{c}{\textbf{Classification}} &
  \multicolumn{5}{c}{\textbf{AA Localization}} \\ 
  \cmidrule(lr){2-5}
  \cmidrule(lr){6-10}
\multicolumn{1}{c}{} &
  \multicolumn{2}{c}{Pneumonia} &
  \multicolumn{2}{c}{Pneumothorax} &
  \multicolumn{5}{c}{Accuracy @ IoU = 0.25}
  \\ 
  \cmidrule(lr){2-5} 
  \cmidrule(lr){6-10}
\multicolumn{1}{c}{} & $\alpha$ & AUPRC & $\alpha$ & AUPRC & LAL& LL& RAL& RLL& RL \\ 
\cmidrule(lr){1-1}
\cmidrule(lr){2-5}
\cmidrule(lr){6-10}
AGXNet w/o AGA      & - & 0.57  & - & 0.55          & 0.02 & 0.10 & 0.08 & 0.43 & 0.12          \\
AGXNet w/ AGA    & - & 0.57  & - & \textbf{0.57} & 0.28 & 0.22 & 0.49 & 0.63 & \textbf{0.22} \\
AGXNet w/ AGA + PU &
  0.14 &
  \textbf{0.62} &
  0.03 &
  \textbf{0.57} &
  \textbf{0.38} &
  \textbf{0.23} &
  \textbf{0.50} &
  \textbf{0.69} &
  \textbf{0.22} \\ 
  \bottomrule
\end{tabular}
\end{adjustbox}
\label{table:2}
\end{table*}

\begin{figure}
\centering
\includegraphics[width=0.7\textwidth]{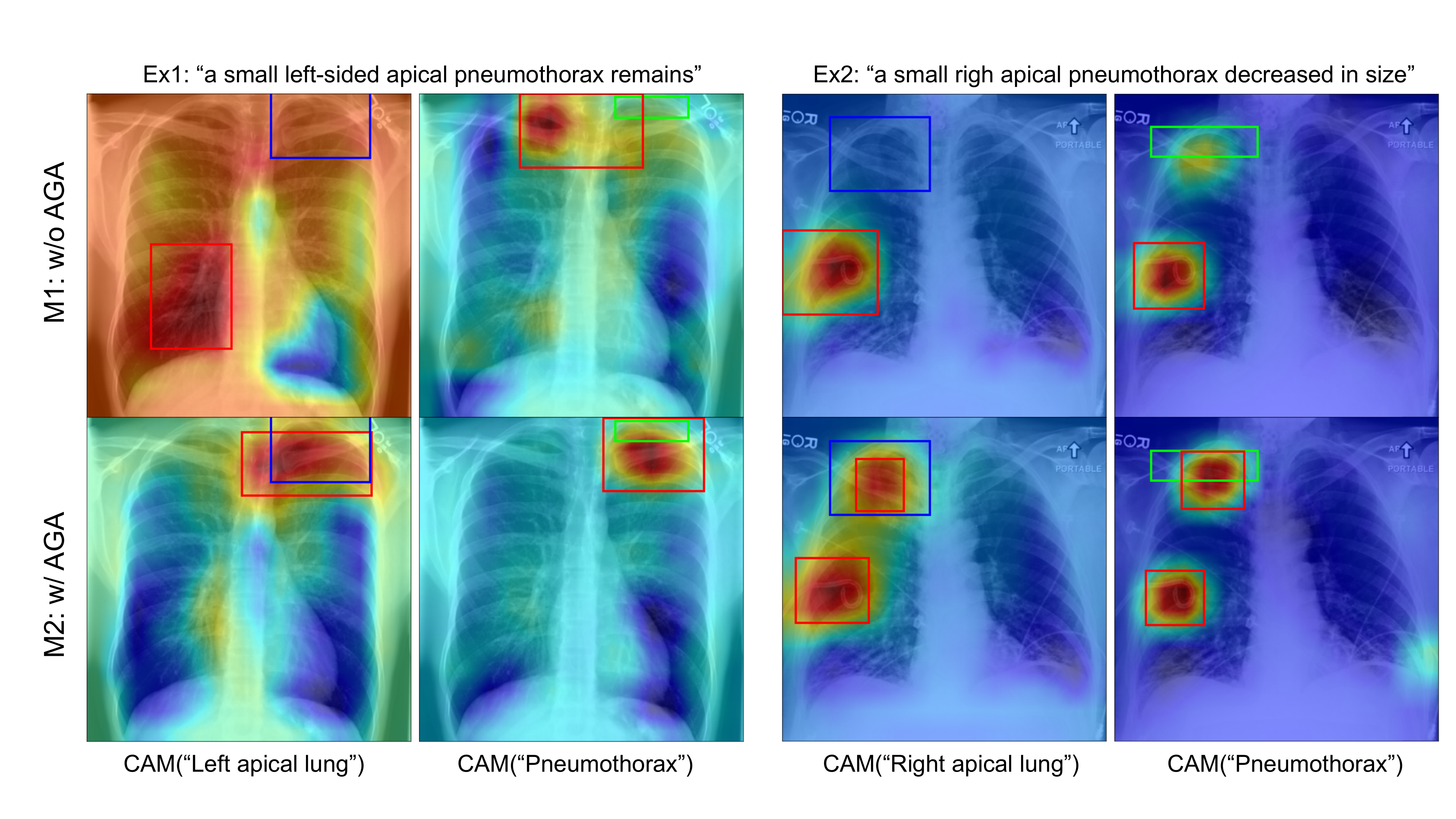}
\caption{AGXNet with (M2) or without (M1) the AGA module for pneumothorax detection. Heatmaps indicate the CAMs. The blue and green boxes stand for ground truth anatomy and disease annotations. The red boxes are generated from the CAMs.
}
\label{fig:3}
\end{figure}

\noindent \textbf{Ablation Studies.} Tab.~\ref{table:2} shows additional results of ablation studies for AGXNet, including (1) disease classification performance on the positive and negative samples in the test set using the area under the precision-recall curve (AUPRC), and (2) anatomical abnormality localization accuracy at $T(\text{IoU})=0.25$ for $\{a_j\}$, where $A(a_j, o_k)=P$ and $o_k \in \{\text{pneumonia, pneumothorax}\}$. \textbf{Effect of AGA:} Results in Tab.~\ref{table:1} and Tab.~\ref{table:2} show that the AGA module significantly improved the results of pneumothorax detection and anatomical abnormality detection, suggesting that the attention module mutually enhanced both types of features. Fig.~\ref{fig:3} shows the qualitative comparison of two models with (M2) and without (M1) the AGA module. M2 correctly detects pneumothorax and the relevant abnormal anatomical landmarks in both examples, while M1 fails to do so. Furthermore, in the second example, a pigtail catheter is applied to treat pneumothorax and acts as the only discriminative feature used by M1, while M2 is more robust against this shortcut and detects disease in the correct anatomical location. \textbf{Effect of PU Learning:} Tab.~\ref{table:2} shows that the estimated mixture proportion of positive examples ($\alpha$) is significantly higher in pneumonia ($14\%$) than in pneumothorax ($3\%$), reflecting the fact that there is considerable variability in pneumonia diagnoses~\cite{neuman2012variability}. Accordingly, results in Tab.~\ref{table:1} and Tab.~\ref{table:2} show that using the PU learning technique improves the performance of pneumonia classification and localization, while it is less effective for pneumothorax whose label noise is intrinsically low. 

\begin{table*}[h]
\caption{The test AUROCs for the NIH Chest X-ray disease classification task.}
\begin{adjustbox}{max width=0.9\textwidth, center}
\begin{tabular}{cccccccccc}
\toprule
\textbf{Method} &
  \scriptsize{Atelectasis} &
  \scriptsize{Cardiomegaly} &
  \scriptsize{Effusion} &
  \scriptsize{Infiltration} &
  \scriptsize{Mass} &
  \scriptsize{Nodule} &
  \scriptsize{Pneumonia} &
  \scriptsize{Pneumothorax} &
  Mean \\
\midrule
Scratch & 0.79 & 0.91 & 0.88 & 0.69 & 0.81 & 0.70 & 0.70 & 0.82          & 0.79 \\
Wang et al.~\cite{wang2017chestx}     & 0.72 & 0.81 & 0.78 & 0.61 & 0.71 & 0.67 & 0.63 & 0.81          & 0.72 \\
Wang et al.~\cite{wang2018tienet}      & 0.73 & 0.84 & 0.79 & 0.67 & 0.73 & 0.69 & 0.72 & 0.85          & 0.75 \\
Liu et al.~\cite{liu2019align}      & 0.79 & 0.91 & 0.88 & 0.69 & 0.81 & 0.73 & 0.75 & 0.89          & 0.80 \\
Rajpurkar et al.~\cite{rajpurkar2017chexnet} & 0.82 & 0.91 & 0.88 & \textbf{0.72} & 0.86 & 0.78 & 0.76 & 0.89 & 0.83 \\ 
Han et al.~\cite{han2022knowledge} & 0.84 & \textbf{0.93} & 0.88 & \textbf{0.72} & \textbf{0.87} & 0.79 & \textbf{0.77}  & \textbf{0.90} & \textbf{0.84} \\ 
\midrule
AGXNet - Ana.      & 0.84 & 0.91 & \textbf{0.90} & 0.71 & 0.86 & \textbf{0.80} & 0.74 & 0.86          & 0.83 \\
AGXNet - Obs.       & \textbf{0.85} & 0.92 & \textbf{0.90} & \textbf{0.72} & 0.86 & \textbf{0.80} & 0.76 & 0.87          & \textbf{0.84} \\
AGXNet - Both &
  \textbf{0.85} &
  0.92 &
  \textbf{0.90} &
  \textbf{0.72} &
  \textbf{0.87} &
  \textbf{0.80} &
  0.76 &
  0.88 &
  \textbf{0.84} \\
\bottomrule
\end{tabular}
\end{adjustbox}
\label{table:3}
\end{table*}

\begin{table*}[h]
\caption{The localization accuracy for the NIH Chest X-ray disease localization task. The highest values are highlighted in bold, and the best results without BBox annotation are underlined.}
\begin{adjustbox}{max width=0.9\textwidth, center}
\begin{tabular}{cccccccccccc}
\toprule
  \textbf{T(IoU)}&
  \textbf{Method} &
  \textbf{Supervised} &
  \scriptsize{Atelectasis} &
  \scriptsize{Cardiomegaly} &
  \scriptsize{Effusion} &
  \scriptsize{Infiltration} &
  \scriptsize{Mass} &
  \scriptsize{Nodule} &
  \scriptsize{Pneumonia} &
  \scriptsize{Pneumothorax} &
  Mean \\ 
  \midrule
\multirow{5}{*}{0.1} 
&
  Scratch &
  \xmark &
  0.46 &
  0.85 &
  0.61 &
  0.40 &
  0.48 &
  0.11 &
  0.56 &
  0.25 &
  0.47\\
&
  Han et al.~\cite{han2022knowledge} &
  \cmark &
  \textbf{0.72} &
  0.96 &
  \textbf{0.88} &
  \textbf{0.93} &
  \textbf{0.74} &
  \textbf{0.45} &
  0.65 &
  \textbf{0.64} &
  \textbf{0.75} \\
&
  Wang et al.~\cite{wang2017chestx} &
  \xmark &
  0.69 &
  0.94 &
  0.66 &
  \underline{0.71} &
  0.40 &
  0.14 &
  0.63 &
  0.38 &
  0.57 \\
 &
  AGXNet - Ana. &
  \xmark &
  0.64 &
  0.99 &
  0.66 &
  0.69 &
  \underline{0.69} &
  \underline{0.29} &
  \underline{\textbf{0.73}} &
  0.29 &
  0.62 \\
 &
  AGXNet - Obs. &
  \xmark &
  0.61 &
  0.99 &
  \underline{0.70} &
  0.70 &
  \underline{0.69} &
  0.23 &
  0.72 &
  0.55 &
  0.65 \\
 & AGXNet - Both &
 \xmark &
 \underline{0.71} & 
 \underline{\textbf{1.00}} & 
 0.69 & 
 0.68 & 
 0.68 & 
 0.22 & 
 0.71 & 
 \underline{0.62} &
 \underline{0.66} \\
 
\midrule
\multirow{5}{*}{0.3} 
&
  Scratch &
  \xmark &
  0.10 &
  0.45 &
  0.34 &
  0.15 &
  0.14 &
  0.00 &
  0.31 &
  0.06 &
  0.19 \\
&
  Han et al.~\cite{han2022knowledge} &
  \cmark &
  \textbf{0.39} &
  \textbf{0.85} &
  \textbf{0.60} &
  \textbf{0.67} &
  \textbf{0.43} &
  \textbf{0.21} &
  0.40 &
  \textbf{0.45} &
  \textbf{0.50} \\
&
  Wang et al.~\cite{wang2017chestx} &
  \xmark &
  0.24 &
  0.46 &
  0.30 &
  0.28 &
  0.15 &
  \underline{0.04} &
  0.17 &
  0.13 &
  0.22 \\
 &
  AGXNet - Ana. &
  \xmark &
  0.20 &
  0.48 &
  0.40 &
  0.43 &
  \underline{0.28} &
  0.00 &
  0.46 &
  0.13 &
  0.30 \\
 &
  AGXNet - Obs. &
  \xmark & 
  0.22 &
  \underline{0.62} &
  0.45 &
  \underline{0.46} &
  0.27 &
  0.01 &
  \underline{\textbf{0.54}} &
  0.31 &
  \underline{0.36} \\
 & AGXNet - Both &
 \xmark &
 \underline{0.27} &
 0.56 &
 \underline{0.47} &
 0.45 &
 0.27 &
 0.01 &
 0.50 &
 \underline{0.36} &
 \underline{0.36} \\

 \bottomrule
\end{tabular}
\label{table:4}
\end{adjustbox}
\end{table*}

\noindent \textbf{Transfer Learning on NIH Chest X-ray.} We pre-trained an $\text{AGXNet}+\text{PU}$ model on the MIMIC-CXR dataset using the 46 anatomy labels and 8 observation labels listed in C.1. We then fine-tuned the encoder(s) and re-trained a classifier using the NIH Chest X-ray dataset~\cite{wang2017chestx}. We investigated three variants of fine-tuning regimes: (1) only using anatomy encoder, (2) only using observation encoder, and (3) using both encoders and their concatenated embeddings. We compare our transferred models with a baseline model trained from scratch using the NIH Chest X-ray dataset and a series of relevant baselines. We evaluate the classification performance using the area under the ROC curve (AUROC) score and the localization accuracy at $T(\text{IoU}) = 0.1$, $0.3$ across the 8 diseases. Tab.~\ref{table:3} shows that all variants of fine-tuned AGXNet models achieve disease classification performances comparable to the SOTA method~\cite{han2022knowledge}, demonstrating that both learned anatomy and observation features are robust and transferable. Tab.~\ref{table:4} shows that all variants of fine-tuned AGXNet models outperform learning from scratch and the existing CAM-based baseline method~\cite{wang2017chestx} in disease localization. Note that the model proposed by Han et al.~\cite{han2022knowledge} achieved higher accuracy by utilizing the ground truth BBox during training, therefore it is not directly comparable and should be viewed as an upper bound method.  
\label{sec:experiment}

\section{Conclusion}
In this work, we propose a novel WSL framework to incorporate anatomical contexts mentioned in radiology reports to facilitate disease detection on corresponding CXR images. In addition, we use a PU learning approach to explicitly handle noise in unlabeled data. Experimental evaluations on the MIMIC-CXR dataset show that the addition of anatomic knowledge and the use of PU learning improve abnormality localization. Experiments on the NIH Chest X-ray datasets demonstrate that the learned anatomical and pathological features are transferable and encode robust classification and localization information.
\label{sec:conclusion}

\subsubsection{Acknowledgements} This work was partially supported by NIH Award Number 1R01HL141813-01, NSF 1839332 Tripod+X, and SAP SE.
We are grateful for the computational resources provided by Pittsburgh SuperComputing grant number TG-ASC170024.

%
%
%
\bibliographystyle{splncs04}
\bibliography{paper1726}

\appendix
\setcounter{table}{0}
\setcounter{figure}{0}
\renewcommand\thefigure{\thesection.\arabic{figure}}    
\renewcommand\thetable{\thesection.\arabic{table}}  
\newpage
\section{Example of Adjacency Matrix}
\begin{figure}[ht]
\centering\includegraphics[width=1\textwidth]{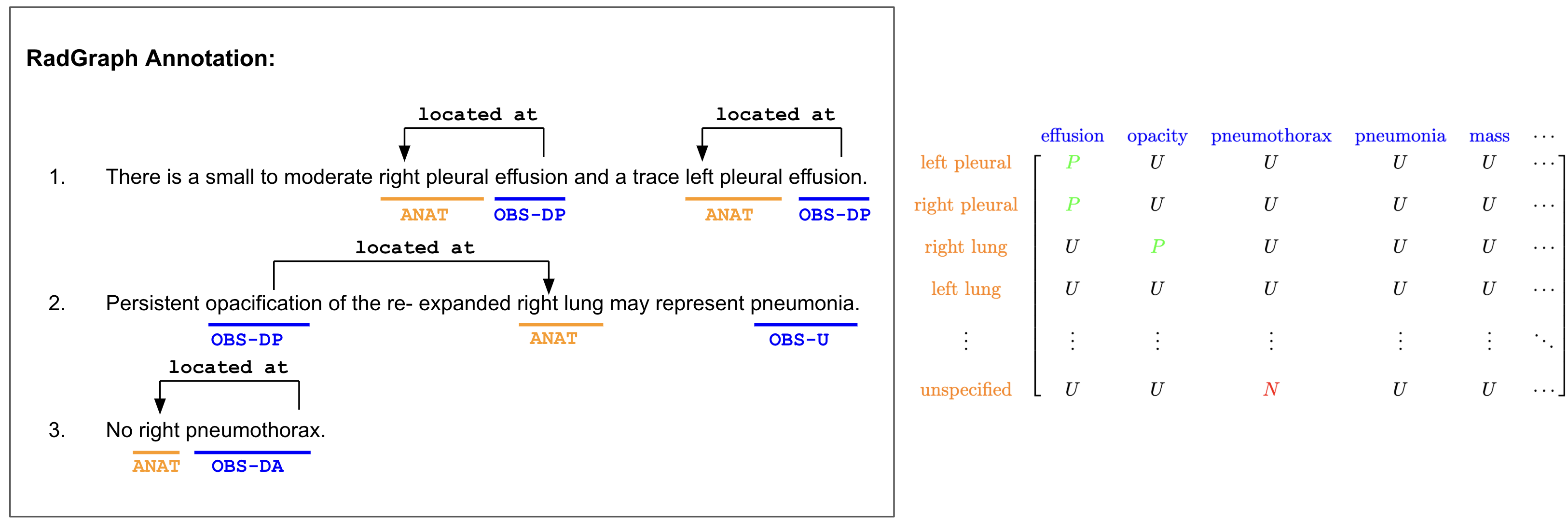}
\caption{An example of adjacency matrix derived from RadGraph annotations for a report. ANAT denotes anatomy mention. OBS-DP, OBS-DA and OBS-U denote observation mentions that are definitely present, definitely absent and uncertain.}
\label{fig:1}
\end{figure}

\section{Anatomical Abnormality Localization}
\begin{figure}[h]
\centering
\includegraphics[width=0.8\textwidth]{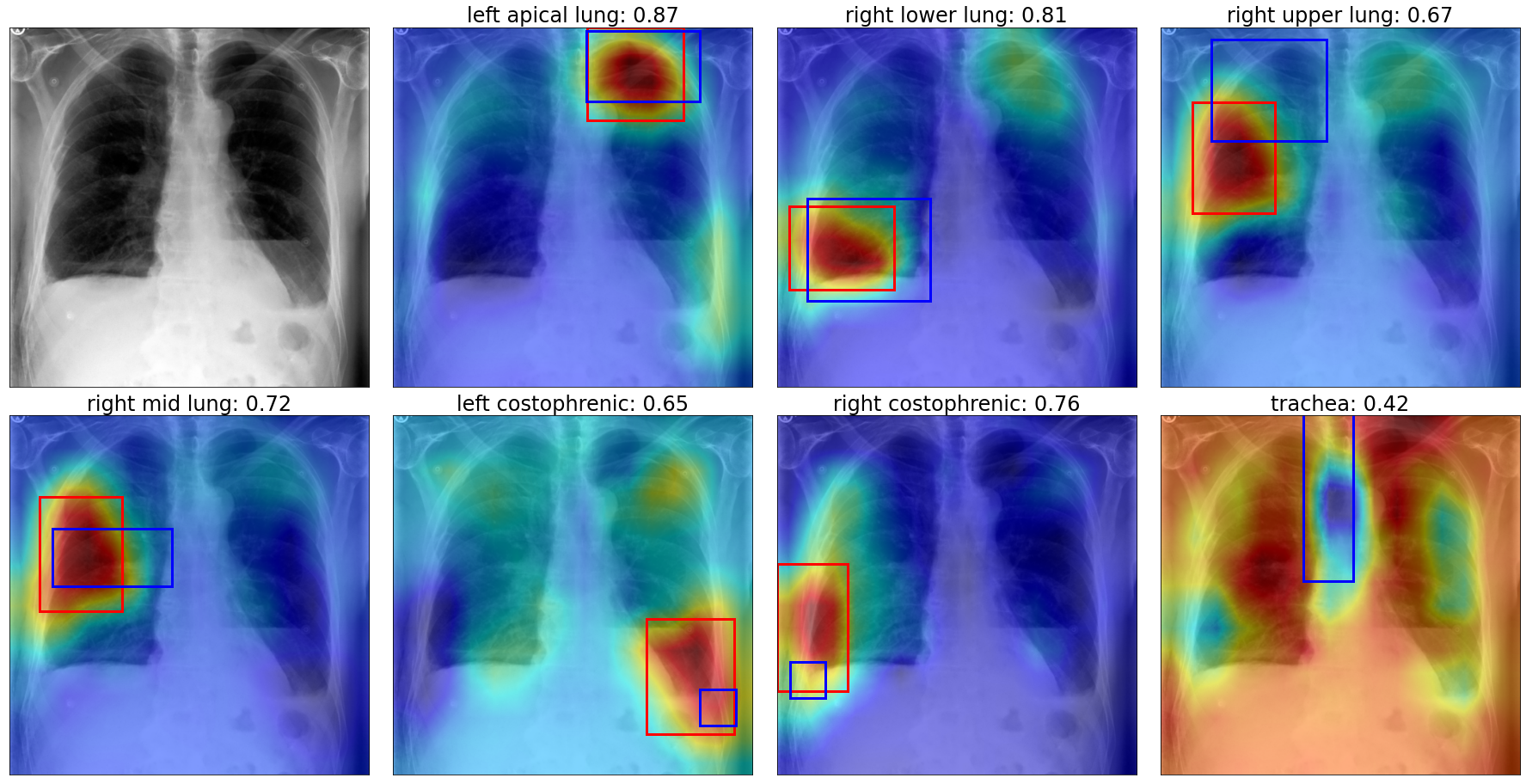}
\caption{Abnormality localization for different anatomical landmarks within the same image. The blue boxes stand for ground truth anatomy annotations. The red boxes stand for the detected abnormal regions in the anatomical landmark. The predicted probabilities of anatomy network in AGXNet are shown in the titles. 
}
\label{fig:4}
\end{figure}

\newpage
\section{Transfer Learning on the NIH Chest X-ray Dataset}
\begin{table*}
\caption{Summary of observation labels used for pre-training.}
\centering
\begin{adjustbox}{max width=0.9\textwidth, center}
\begin{tabular}{lccclll}
\toprule 
\addlinespace
\textbf{Observation} & \textbf{P (\%)} & \textbf{N (\%)} & \textbf{U (\%)} & \multicolumn{3}{c}{\textbf{Top 3 Associated Anatomical Landmarks (\% in P)}} \\ 
\addlinespace
\cmidrule(lr){1-1}
\cmidrule(lr){2-4}
\cmidrule(lr){5-7}
Atelectasis   & 21.9 & 0.3  & 77.8 &  Lung bases (28.4)        & Unspecified (15.6)      & Left lower lung (12.5) \\
Effusion      & 23.6 & 49.2 & 27.2 & Pleural (40.5)           & Right pleural (22.5)    & Left pleural (21.2)    \\
Pneumonia     & 3.5  & 14.1 & 82.4 & Unspecified (68.6)       & Lower left lobe (7.5)   & Lower right lobe (6.7) \\
Pneumothorax \hspace{5px}  & 3.8  & 60.8 & 35.5 & Right apical lung (22.1) & Left apical lung (16.5) & Right lung (15.9)      \\
Cardiomegaly  & 14.1 & 0.2  & 85.7 & Unspecified (99.8)       & Mediastinal (0.05)      & Heart (0.02)           \\
Enlarge       & 13.6 & 0.8  & 85.6 & Heart (77.1)             & Mediastinal (4.7)       & Unspecified (2.9)      \\
Nodule        & 1.6  & 0.6  & 97.8 & Unspecified (13)         & Upper right lobe (5.4)  & Upper left lobe (4)    \\
Mass          & 1.1  & 0.4  & 98.5 & Unspecified (11.4)       & Right hilar (10.1)      & Mediastinal (7.5)      \\ 
\bottomrule
\end{tabular}
\end{adjustbox}
\label{table:5}
\end{table*}

\begin{table*}[h]
\caption{The localization accuracy for the NIH Chest X-ray disease localization task by IoU levels.}
\begin{adjustbox}{max width=0.9\textwidth, center}
\begin{tabular}{ccccccccccc}
\toprule
  \textbf{T(IoU)}&
  \textbf{Method} &
  \scriptsize{Atelectasis} &
  \scriptsize{Cardiomegaly} &
  \scriptsize{Effusion} &
  \scriptsize{Infiltration} &
  \scriptsize{Mass} &
  \scriptsize{Nodule} &
  \scriptsize{Pneumonia} &
  \scriptsize{Pneumothorax} &
  Mean \\ 
  \midrule
\multirow{3}{*}{0.1} 
 &
  AGXNet - Ana. &
  0.64 &
  0.99 &
  0.66 &
  0.69 &
  0.69 &
  0.29 &
  0.73 &
  0.29 &
  0.62 \\
 &
  AGXNet - Obs. &
  0.61 &
  0.99 &
  0.70 &
  0.70 &
  0.69 &
  0.23 &
  0.72 &
  0.55 &
  0.65 \\
 & AGXNet - Both &
 0.71 & 
 1.00 & 
 0.69 & 
 0.68 & 
 0.68 & 
 0.22 & 
 0.71 & 
 0.62 &
 0.66 \\
 
\midrule
\multirow{3}{*}{0.3} 
 &
  AGXNet - Ana. &
  0.20 &
  0.48 &
  0.40 &
  0.43 &
  0.28 &
  0.00 &
  0.46 &
  0.13 &
  0.30 \\
 &
  AGXNet - Obs. &
  0.22 &
  0.62 &
  0.45 &
  0.46 &
  0.27 &
  0.01 &
  0.54 &
  0.31 &
  0.36 \\
 & AGXNet - Both &
 0.27 &
 0.56 &
 0.47 &
 0.45 &
 0.27 &
 0.01 &
 0.50 &
 0.36 &
 0.36 \\
 
 \midrule
\multirow{3}{*}{0.5} 
&
  AGXNet - Ana. &
  0.03 &
  0.03 &
  0.10 &
  0.21 &
  0.09 &
  0.00 &
  0.19 &
  0.02 &
  0.08 \\
 &
  AGXNet - Obs. &
  0.03 &
  0.01 &
  0.13 &
  0.24 &
  0.14 &
  0.00 &
  0.26 &
  0.08 &
  0.11 \\
 & AGXNet - Both &
 0.03 &
 0.01 &
 0.12 &
 0.26 &
 0.11 &
 0.00 & 
 0.25 &
 0.14 &
 0.12 \\
 \bottomrule
\end{tabular}
\label{table:7}
\end{adjustbox}
\end{table*}
\label{sec:supplemental}

\section{Effect of Scaling Hyperparameter $\beta$ in Residual Attention}
\begin{table*}
\caption{Disease localization on MIMIC-CXR.}
\begin{adjustbox}{max width=0.8\textwidth, center}
\begin{tabular}{lccccccc}
\toprule
\multicolumn{1}{c}{} &
  \multicolumn{1}{c}{} &
  \multicolumn{2}{c}{\textbf{IoU @ 0.1}} &
  \multicolumn{2}{c}{\textbf{IoU @ 0.25}} &
  \multicolumn{2}{c}{\textbf{IoU @ 0.5}} \\ 
  \cmidrule(lr){3-4}
  \cmidrule(lr){5-6}
  \cmidrule(lr){7-8}
\multicolumn{1}{l}{\multirow{-2}{*}{\textbf{Disease}}} &
  \multicolumn{1}{c}{\multirow{-2}{*}{\textbf{AGXNet}}} &
  \multicolumn{1}{c}{\scriptsize Recall} &
  \multicolumn{1}{c}{\scriptsize Precision} &
  \multicolumn{1}{c}{\scriptsize Recall} &
  \multicolumn{1}{c}{\scriptsize Precision} &
  \multicolumn{1}{c}{\scriptsize Recall} &
  \multicolumn{1}{c}{\scriptsize Precision} \\ 
  \midrule
 &
  $\beta=0$ &
  $0.67_{\pm 0.04}$ &
  $0.41_{\pm 0.03}$ &
  $0.54_{\pm 0.04}$ &
  $0.33_{\pm 0.03}$ &
  $\bm{0.16_{\pm 0.02}}$ &
  $\bm{0.10_{\pm 0.01}}$ \\
 &
  $\beta=0.05$ &
  $0.71_{\pm 0.02}$ &
  $0.41_{\pm 0.02}$ &
  $0.58_{\pm 0.02}$ &
  $0.33_{\pm 0.01}$ &
  $0.15_{\pm 0.01}$ &
  $0.09_{\pm 0.00}$ \\
\multirow{-3}{*}{Pneumothorax } &
  $\beta=0.1$ &
  $\bm{0.74_{\pm 0.03}}$ &
  $\bm{0.43_{\pm 0.02}}$ &
  $\bm{0.58_{\pm 0.03}}$ &
  $\bm{0.34_{\pm 0.02}}$ &
  $0.13_{\pm 0.01}$ &
  $0.07_{\pm 0.01}$ \\
  \midrule
 &
  $\beta=0$ &
  $0.69_{\pm 0.01}$ &
  $0.44_{\pm 0.01}$ &
  $0.59_{\pm 0.02}$ &
  $0.37_{\pm 0.02}$ &
  $0.18_{\pm 0.02}$ &
  $0.11_{\pm 0.01}$ \\
 &
  $\beta=0.05$ &
  $0.71_{\pm 0.01}$ &
  $0.45_{\pm 0.02}$ &
  $0.58_{\pm 0.02}$ &
  $0.36_{\pm 0.02}$ &
  $0.17_{\pm 0.01}$ &
  $0.11_{\pm 0.01}$ \\
\multirow{-3}{*}{\footnotesize Pneumonia} &
  $\beta=0.1$ &
  $\bm{0.73_{\pm 0.01}}$ &
  $\bm{0.47_{\pm 0.01}}$ &
  $\bm{0.62_{\pm 0.01}}$ &
  $\bm{0.40_{\pm 0.01}}$ &
  $\bm{0.20_{\pm 0.01}}$ &
  $\bm{0.13_{\pm 0.01}}$ \\ 
  \bottomrule
\end{tabular}
\end{adjustbox}
\label{table:6}
\end{table*}

\end{document}